\documentclass[12pt]{article}
\usepackage[letterpaper,top=2cm,bottom=2cm,left=3cm,right=3cm,marginparwidth=1.75cm]{geometry}

\usepackage{eccvabbrv}

\usepackage{graphicx}
\usepackage{booktabs}
\usepackage{multirow}
\usepackage{wrapfig}

\usepackage[accsupp]{axessibility}  

\usepackage[pagebackref,breaklinks,colorlinks]{hyperref}

\newcommand{\framework}{OODSmoother\xspace}
\newcommand{\defense}{SemPrior\xspace}
\usepackage{mathtools}
\usepackage{diagbox}

\DeclarePairedDelimiter\floor{\lfloor}{\rfloor}

\begin{document}

\title{Defending Object Detectors against Patch Attacks with Out-of-Distribution Smoothing} 
\date{}


\author{\Large{Ryan Feng${}^1$, Neal Mangaokar${}^1$, Jihye Choi${}^2$},\\ \Large{Somesh Jha${}^2$, Atul Prakash${}^1$}\vspace{0.1em}\\
\normalsize{${}^1$University of Michigan, ${}^2$University of Wisconsin}\vspace{0.1em}\\
\normalsize{${}^1$\{rtfeng, nealmgkr, aprakash\}@umich.edu, ${}^2$\{jihye, jha\}@cs.wisc.edu
}}

\maketitle

\begin{abstract}
  Patch attacks against object detectors have been of recent interest due to their being physically realizable and more closely aligned with practical systems. In response to this threat, many new defenses have been proposed that train a patch segmenter model to detect and remove the patch before the image is passed to the downstream model. We unify these approaches with a flexible framework, \framework, which characterizes the properties of approaches that aim to remove adversarial patches. 
   This framework naturally guides us to design 1) a novel adaptive attack that breaks existing patch attack defenses on object detectors, and 2) a novel defense approach \defense that takes advantage of semantic priors. 
  Our key insight behind \defense is that the existing machine learning-based patch detectors struggle to learn semantic priors and that explicitly incorporating them can improve performance. We find that \defense alone provides up to a 40\% gain, or up to a 60\% gain when combined with existing defenses.

\end{abstract}

\section{Introduction}
Machine learning models today remain vulnerable to adversarial examples~\cite{goodfellow2014explaining,szegedy2013intriguing,athalye2018obfuscated,athalye2018synthesizing,eykholt2018robust,feng2022graphite,tramer2020adaptive}, where perturbed inputs lead to unexpected model outputs. Such adversarial examples take a variety of forms, including digital attacks~\cite{goodfellow2014explaining,szegedy2013intriguing} and physical~\cite{eykholt2018robust,athalye2018synthesizing,feng2022graphite} attacks, where the attack can be physically-realized in the real-world in the form of printed stickers~\cite{eykholt2018robust,feng2022graphite} or 3D objects~\cite{athalye2018synthesizing}. Thus, the patch attack has been of increasing interest due to its ability to practically inject an attack via the insertion of a printed physical patch into the scene.

A variety of patch attacks defenses have thus been proposed, including several certified~\cite{levine2020randomized,chiang2020certified,xiang2022patchcleanser,xiang2021patchguard,xiang2023objectseeker,lin2021certified} and empirical~\cite{xu2023patchzero,liu2022segment,liang2021we,zhou2020information,tarchoun2023jedi,chen2023jujutsu} defenses, with many of these defenses designed around the operation of identifying and then removing the patch. Such defenses rely on being able to accurately identify the patch attack without false positives and remove the effects of identified patches with a variety of techniques, including blacking them out~\cite{liu2022segment} or setting it to the image's mean color~\cite{xu2023patchzero}.

\begin{figure}[tb]
  \centering
  \includegraphics[height=3.2cm]{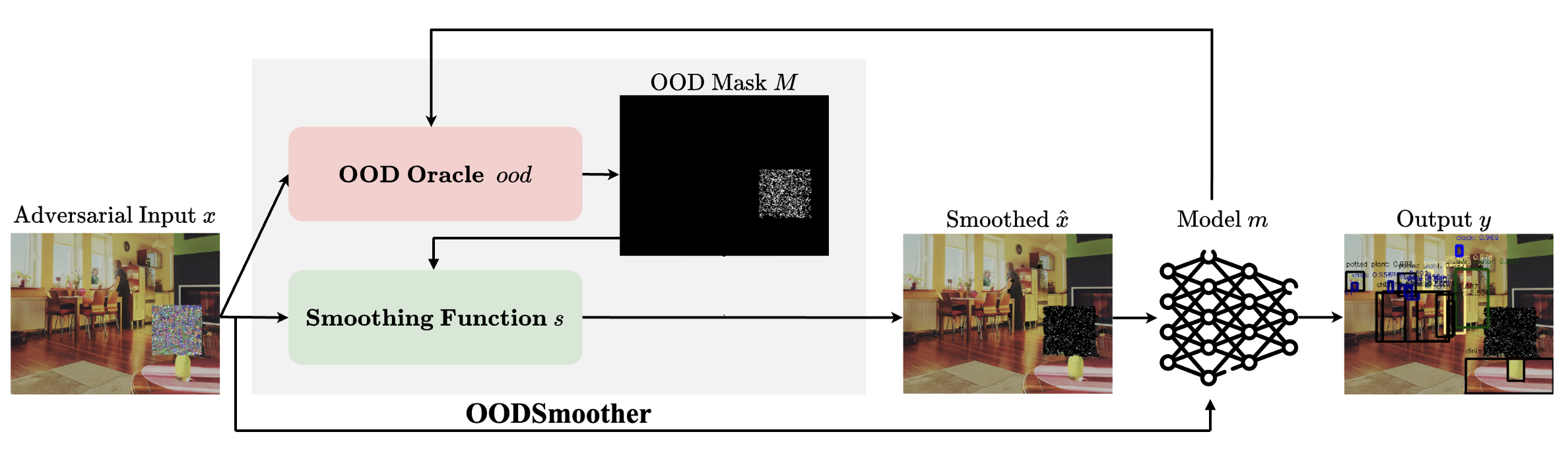}
  \caption{Our \framework framework consists of two components: an OOD oracle $ood$ and a OOD smoother $s$. The OOD oracle takes in the input $x$ and the model $m$ and provides OOD scores that are then smoothed by the OOD smoother $s$ before being passed into the downstream model.
  }
  \label{fig:framework}
  \vspace{-0.5cm}
\end{figure}

Our first key contribution is that we unify these types of defenses under a \emph{general framework called \framework} (Section~\ref{framework}), as shown in Fig.~\ref{fig:framework}. 
The key insight of our framework is that patch segmenter and removal systems can be viewed generally through the lens of out-of-distribution (OOD) feature detection and OOD smoothing. Our framework thus consists of two components: 1) an OOD oracle and 2) an OOD smoother. 
The OOD oracle's job is to score how input features are far from natural features, according to some metric (e.g., a patch segmented). 
Ideally, the OOD oracle would perfectly separate natural and adversarial features. The OOD smoother's job is then to, given the OOD oracle, smooth the image to remove the OOD features (e.g., a patch removal). 

As a result of our \framework framework, we make our second key contribution: a novel adaptive attack that \emph{breaks existing object detection defenses against patch attacks} (Section~\ref{existing}). 
Our key insight is that simultaneously attacking the OOD oracle and the downstream model is much more effective than simply attacking the end-to-end system. We evaluate the efficacy of our attacks against two SOTA defenses, SAC~\cite{liu2022segment} and PatchZero~\cite{xu2023patchzero}, reducing the mAP @ IoU = 0.5 from 53.6\% to 6.2\% on SAC and MS COCO~\cite{lin2014microsoft} and from 68.9\% to 7.0\% on PatchZero and Pascal VOC~\cite{everingham2010pascal} with $100 \times 100$ patches.

Finally, our third key contribution is a novel defense, \defense, that \emph{incorporates semantic priors} (Section~\ref{ours}). The defense is inspired by two insights: (1) with our flexible framework \framework, we can simply swap out the OOD oracle with a set of OOD oracles, e.g., that label anomalous color patterns so as to help restrict the attack space; and (2) existing ML based patch detectors such as those in SAC~\cite{liu2022segment} or PatchZero~\cite{xu2023patchzero} struggle to learn simple semantic priors based on color properties without explicit supervision.  We find that on adaptive attacks, \defense increases the adversarial mAP by up to 40\%. We also find that composing \defense's oracles with SAC's is particularly effective, increasing the adversarial mAP by up to 60\% over just SAC.

\section{Related Work}
\subsection{Patch Attacks} The adversarial patch attack was originally proposed by Brown \etal~\cite{brown2017adversarial} as a universal attack that could be printed and placed in a scene to cause misclassification in the physical world. Since then, other patch attacks have come out such as LaVAN~\cite{karmon2018lavan} for small digital patch attacks, GRAPHITE~\cite{feng2022graphite} for hard-label physical attacks, DPatch~\cite{liu2018dpatch} for object detection, and an attack against multi-modal systems with RGB and infrared cameras~\cite{wei2023unified}. However, these works largely are not concerned about adaptively attacking defenses that include a patch removal system.

\subsection{Patch Attack Defenses} A variety of defenses against patch attacks have been proposed, including early approaches on digital watermarking~\cite{hayes2018visible} or gradient smoothing~\cite{naseer2019local}, certified defenses~\cite{levine2020randomized,chiang2020certified,xiang2022patchcleanser,xiang2021patchguard,xiang2023objectseeker,lin2021certified}, patch removal approaches~\cite{xu2023patchzero,liu2022segment,liang2021we,zhou2020information,tarchoun2023jedi,chen2023jujutsu}, and voting approaches~\cite{xiang2022patchcleanser,xiang2023objectseeker,lin2021certified}. Many of these defenses generally follow a template of identifying OOD pixels that are indicative of the patch and removing them, or voting on image segments to deduce the patch location, motivating our common framework. We focus on two recently proposed SOTA defenses as discussed below.
\begin{enumerate}
    \item \textbf{SAC.} SAC~\cite{liu2022segment} is a defense for object detectors against patch attacks devised as a ``segment'' and ``complete'' strategy to mask out the patch attack pixels. To detect the adversarial patch, the authors apply two steps, with the first consisting of a U-Net~\cite{ronneberger2015u} trained to predict patch pixels as a segmentation problem and the second consisting of a shape completion algorithm to fill out the shape of the pixels predicted by the U-Net. Finally, SAC masks out these pixels, setting them to black before sending them to the downstream detector. To adaptively attack SAC, the authors propose an attack that uses BPDA~\cite{athalye2018obfuscated} with an identity function to approximate the shape completion and thresholding operations.

    \item \textbf{PatchZero.} PatchZero~\cite{xu2023patchzero} is a defense against patch attacks devised to detect the patch and then zero such pixels out. Designed similarly to SAC~\cite{liu2022segment}, PatchZero proposes a two-stage adversarial training loop to train a PSPNet~\cite{zhao2017pyramid} (instead of a U-Net in SAC) to predict the pixels corresponding to the patch as a segmentation problem. Given the outputs of the PSPNet, the authors zero out the region by setting such pixels to the mean color value of the dataset (instead of black in SAC) before giving the image to the downstream model. The authors adaptively attack it with BPDA~\cite{athalye2018obfuscated} with a Sigmoid approximation for the thresholding (instead of identity function). While some minor differences are present, the core idea is much the same, an idea we expand on in Section~\ref{existing} in the context of our \framework framework proposed in Section~\ref{framework}.
\end{enumerate}

\subsection{Adaptive Attacks} Coming up with adaptive attacks to evaluate new defenses remains an open and difficult problem. Common techniques such as BPDA~\cite{athalye2018obfuscated} to overcome gradient obfuscation and AutoAttack~\cite{croce2020reliable} to automatically tune attack hyperparameters are helpful for improving attack evaluation, but can be easily misapplied or be insufficient without thinking critically about the overall system and what loss(es) are being applied~\cite{tramer2020adaptive}. With our proposed framework, we develop a new adaptive attack that is shown in Section~\ref{existing} to be more effective against SAC~\cite{liu2022segment} and PatchZero~\cite{xu2023patchzero} than the originally proposed adaptive attacks by using two separate losses, one to directly attack the patch segmenter and another for attacking the downstream model.
\section{\framework Framework}\label{framework}
We now describe our proposed framework, \framework, a unified and configurable framework that is meant to characterize the set of defense approaches concerned with OOD pixel identification and feature correction. Our key insight with our framework is that many patch attack defenses use some metric or model (i.e., an oracle) to identify OOD pixels and then apply some process to correct for anomalous features (i.e., a smoother). Thus, this framework assists us in analyzing what properties OOD oracles and smoothers should follow and enables us to: 1) develop novel adaptive attacks against such defenses (Section~\ref{existing}), and 2) propose a new defense \defense to introduce semantic priors as OOD oracles (Section~\ref{ours}), based on the discovery that machine learning based patch detectors do not automatically learn such priors.

\subsection{Overview}
The \framework framework consists of two key components: an OOD Oracle $ood$ and an OOD Smoothing Function $s$. The role of the OOD oracle is to provide OOD scores for features given some inherent OOD metric and the model $m$. The role of the OOD smoother $s$ is to smooth images to remove their OOD features. Once the input image $x$ has been smoothed by $s$ to create $\hat{x}$, then $\hat{x}$ is passed into the model $m$. \framework is visualized in Fig.~\ref{fig:framework}.

\subsection{Threat Model}
We assume a patch attack threat model, wherein the adversary will be restricted to modifying a singular small and continuous region of pixels. We assume the adversary has white-box access to the model and defense. The defender is assumed not to have the ability to retrain the model or otherwise modify the architecture or weights. However, the adversary will have the ability to modify these pixels arbitrarily.

\subsection{OOD Oracle}
The job of the OOD oracle $ood(\cdot)$ is to provide OOD scores for input features. An ideal OOD oracle would perfectly separate out features of natural ID images from features introduced in OOD inputs induced from the adversarial patches:
\begin{align*}
    \forall A\in\mathcal{A}, &\forall a_{i,j} \in A, ood(a_{i,j}) \geq \tau \\
    \forall N\in\mathcal{N}, &\forall n_{i,j} \in N, ood(n_{i,j}) \leq \tau,
\end{align*}
where $\mathcal{A}$ refers to a set of OOD patches with misaligned model outputs, $a_{i,j}$ is the pixel of an OOD patch $A$ at location $i,j$, $\mathcal{N}$ is the set of all natural inputs, $n_{i,j}$ is the pixel of a natural input $N$ at location $i,j$, and $\tau$ is some oracle-specific threshold. As an example, the patch segmenter utilized in SAC~\cite{liu2022segment} serves as an OOD oracle for SAC. Note also that, without loss of generality, $ood$ could also be an ensemble of several OOD oracles that work together to make it harder for an OOD input to fool the oracle into thinking it is in-distribution (ID).

\subsection{OOD Smoothing Function}
The goal of the OOD Smoothing Function $s(\cdot)$ is to correct OOD features by bringing them back to ID. An ideal OOD Smoothing Function would obey the property that it finds the minimal change to bring the OOD input back to ID below some OOD threshold for some distance measure $d$:
\begin{equation}
\begin{array}{ccl}
&\text{minimize}_{\hat{x}}&d(\hat{x} - x)\\ 
&\text{subject to}&ood(\hat{x}) < \mu\\
\end{array}
\end{equation}

An ideal OOD Smoothing Function would thus have the following solutions for the optimization:

\begin{enumerate}
    \item If $x$ is a natural input from the distribution at which the model is trained on, then $\hat{x} = x$
    \item If $x$ is an adversarial example derived from some natural input $x_{nat}$, then $\hat{x} = x_{nat}$
    
\end{enumerate}

For patch attacks, 2. is extremely difficult in practice. Since a patch attack is bound only by location, and not by the amount of perturbation applied in the patch, an attacker could arbitrarily set the patch pixels to anything they like at all. Thus, instead of trying to recover the pixels that were originally erased by the patch, in practice, simply trying to remove the patch features is a worthwhile objective~\cite{liu2022segment,xu2023patchzero}.

\subsection{The Attacker's Objective}
The goal of the attacker is then to increase the mAP, which can be done with operations such as introducing fake boxes that do not actually exist (i.e., hallucination), making it such that the object detector $m$ fails to detect real objects (i.e., disappearance), or changing the classification of detected objects (i.e., misclassification). However, in order to bypass the entire pipeline of OOD smoother, the attacker will have to ensure that the attack remains relatively ID to avoid detection by $ood$. We thus formulate the attacker's objective as follows, where $\theta$ refers to the model parameters of the downstream object detector:

\begin{equation}\label{eq:att_obj}
\begin{array}{ccl}
&\text{minimize}_{p \in P}&L(\theta, s(x + p, ood), y)\\ 
&\text{subject to}&ood(x + p) < \mu\\
\end{array}
\end{equation}

\section{Attacking Existing Defenses}\label{existing}
In this section, we now take a step back and, with the benefit of \framework to view OOD smoothing defenses, analyze how best to attack such defenses. We focus on two object detection defenses against patch attacks, SAC~\cite{liu2022segment} and PatchZero~\cite{xu2023patchzero}, and show that our adaptive attack that applies an additional loss function to attack the OOD oracle is much more effective than the original adaptive attacks proposed. This then motivates us to understand why such defenses have these blind spots, and observe that these patch segmenters failed to learn simple distribution shift statistics.

\subsection{Instantiating Existing Defenses}
We focus on two SOTA object detection defenses against patch attacks, SAC~\cite{liu2022segment} and PatchZero~\cite{xu2023patchzero}. In this section, we begin by instantiating these defenses in the \framework framework to help us understand how to adaptively attack these defenses.

SAC~\cite{liu2022segment} consists of a U-Net~\cite{ronneberger2015u} patch segmenter that predicts as a semantic segmentation task the location of the adversarial patch pixels followed by a shape completion algorithm that fills out the mask area and then the image pixels over the resultant mask are set to black. We can thus think of the patch segmenter as the OOD oracle, where the OODness of image features is characterized by what the patch segmenter learned. Then, the OOD smoothing function consists of the shape completion algorithm and masking to black.

PatchZero~\cite{xu2023patchzero} consists of a PSPNet~\cite{zhao2017pyramid} patch segmenter that predicts as a semantic segmentation task the location of the adversarial patch pixels. Then, the image pixels over the mask are set to the mean color of the dataset. PatchZero is trained with a two-stage adversarial training process. Like SAC~\cite{liu2022segment}, we can think of the patch segmenter as the OOD oracle, where the OODness of image features is characterized by what the patch segmenter learned, and the OOD smoothing function as the masking and setting of the pixels to the mean color of the dataset.

In both SAC~\cite{liu2022segment} and PatchZero~\cite{xu2023patchzero}, the authors propose end-to-end BPDA~\cite{athalye2018obfuscated} attacks, with the non-differentiable components estimated by an identity function and by a sigmoid function respectively. Thus, the loss term that they optimize can be formulated as the objective from~\eqref{eq:att_obj}:

$$
L_{e2e} = L(\theta, s(x + p, ood),y)
$$

\begin{figure}[tb]
  \centering
  \includegraphics[height=3.5cm]{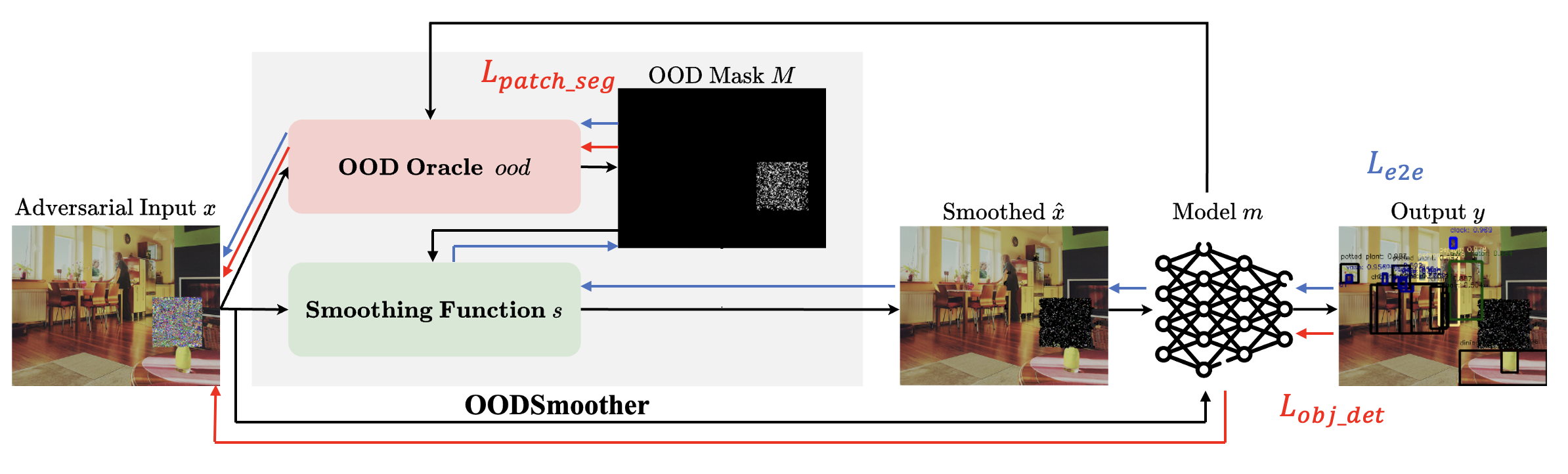}
  \caption{Our adaptive attack uses two loss functions, $L_{patch\_seg}$ and $L_{obj\_det}$, which backprogagates gradients as shown with the red arrows. In the SAC~\cite{liu2022segment} and PatchZero~\cite{xu2023patchzero}defense papers, the authors evaluated their defenses with an attack on the end-to-end pipeline with $L_{e2e}$ and BPDA~\cite{athalye2018obfuscated}. This gradient path is depicted with the blue arrows.
  }
  \label{fig:attack}
\end{figure}

\subsection{Adaptively Attacking an OOD Oracle}

We observe that in the \framework framework, there are two orthogonal tasks: 1) OOD oracle scoring and 2) the downstream model's predictive task. We also observe that for the provided BPDA~\cite{athalye2018obfuscated} adaptive attacks for these defenses, the patch segmenter is still highly accurate for the attacked images. Thus, it raises the question as to whether or not the attack could be focused more on evading the patch segmenter. In particular, we note that the loss function associated with attacking the downstream object detector does not include anything in particular towards evading the OOD oracle and would depend on the BPDA propagating the appropriate gradients.

We thus hypothesize that a more direct approach that separately attacks the two orthogonal tasks with two different loss functions would be more effective. In particular, we keep the loss associated with evading the downstream object detector but then add a second loss term that tries to make the OOD oracle predict every pixel as ID. Formally, we adopt the following optimization problem:
\begin{align}
    \text{maximize}_{p \in \mathcal{P}} ~ & L_{obj\_det} + \lambda L_{patch\_seg} \\ \nonumber
    \text{where} ~ & L_{obj\_det} = L(\theta, x + p, y) \\ \nonumber
    & L_{patch\_seg} = -||ood(x + p)||_2
\end{align}
This formulation essentially takes the attacker's objective in \eqref{eq:att_obj} and approximates the OOD constraint by adding a Lagrangian multiplier in the form of $L_{patch\_seg}$. The pipeline of our adaptive attack vs. the original end-to-end attack is shown in Fig.~\ref{fig:attack}.

\subsection{Evaluating Adaptive Attacks against Existing Defenses}

We include results for attacking existing defenses in Table~\ref{tab:existing_defenses_box} on COCO~\cite{lin2014microsoft} style 101 point interpolated box mAP and mAP @ IoU = 0.5. Following SAC~\cite{liu2022segment}, we evaluate on 1000 images. We test against the authors' attack and parameters, the authors' attack with 1000 steps, and our proposed adaptive attack with 1000 steps. 

For all datasets and all defenses, we find that our proposed adaptive attack is the most effective, bringing the box mAP to below 8\% and the mAP @ IoU = 0.5 to below 12.5\%. We also find that for both metrics on SAC~\cite{liu2022segment}, attacking with 1000 steps on the authors' attack is more effective than their original configuration of 200 steps. For PatchZero~\cite{xu2023patchzero}, we find that the efficacy is about the same regardless of the step configurations. 

\begin{figure}[tb]
  \centering
  \includegraphics[height=8.5cm]{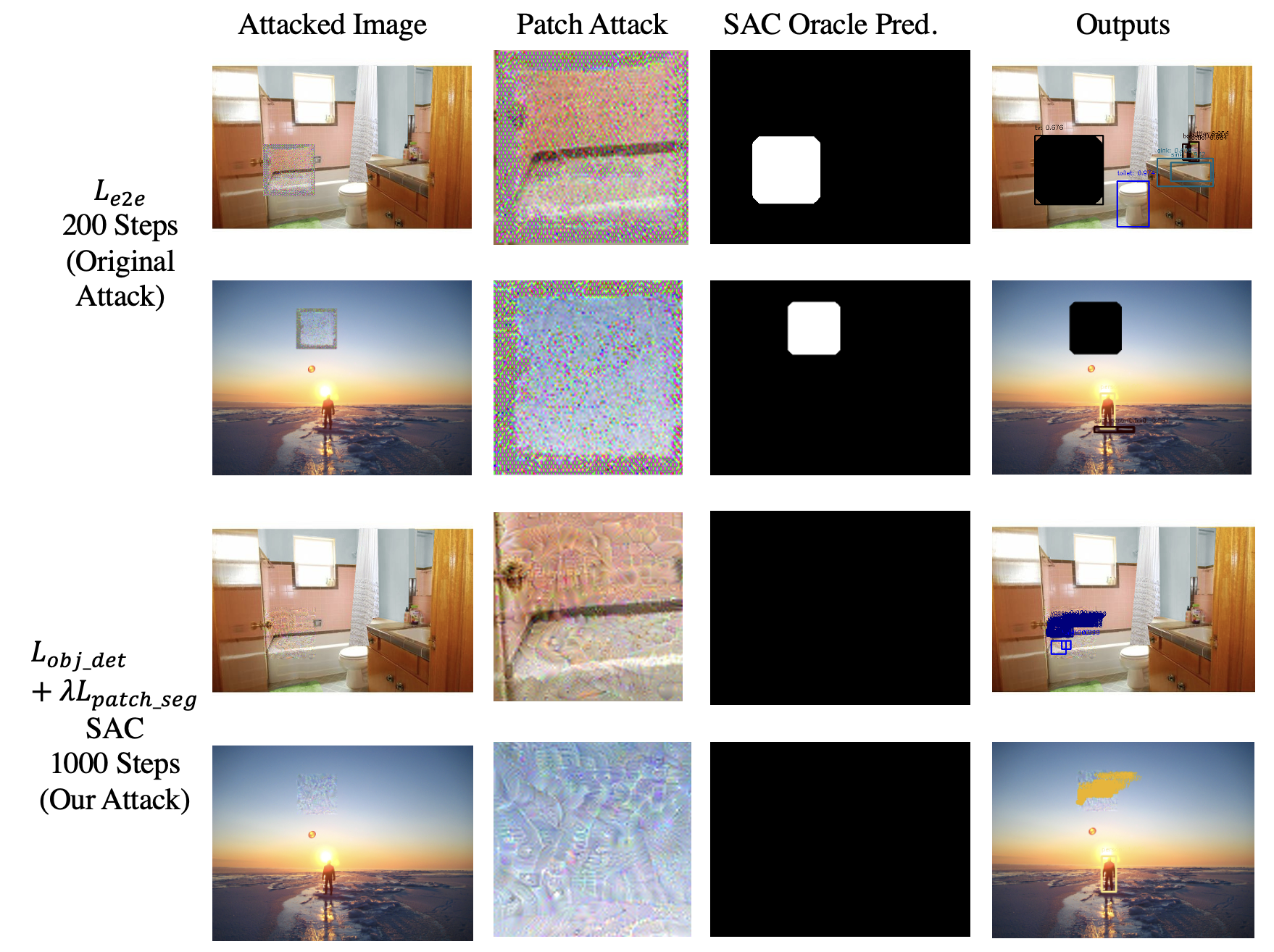}
  \caption{We compare two different attacks against the SAC~\cite{liu2022segment} defense, with the top two rows being the BPDA~\cite{athalye2018obfuscated} the authors used in their evaluation ($L_{e2e}$ with 200 steps) and the bottom two rows being our attack ($L_{obj\_det} + \lambda L_{patch\_seg}$ with 1000 steps). In the top two examples, SAC can successfully mask out the patch. On the bottom two rows, our attack is still able to inject some human-detectable colors in the patch such as neon greens, pinks, and purples that go undetected by SAC. 
  }
  \label{fig:analysis}
\end{figure}

We include some example attacks of the authors' original attack ($L_{e2e}$, 200 steps) and our adaptive attack ($L_{patch\_seg} + \lambda L_{obj\_det}$, 1000 steps) in Fig.~\ref{fig:analysis}. We find that while SAC~\cite{liu2022segment} forces the attack distribution to shift in changing from the original attack to our attack, our adaptive method is still able to evade SAC while including some human-detectable colors in the patch (e.g., neon greens, pinks, purples). This insight then guides us in the design of \defense, to see if incorporating semantic priors can help reduce the attack surface.

\begin{table}[]
\centering
\caption{Evaluating SAC~\cite{liu2022segment} and PatchZero~\cite{xu2023patchzero} on COCO~\cite{lin2014microsoft} style box mAP and mAP @ IoU = 0.5 against adversarial patch attacks over patch sizes of $ 75 \times 75$, $ 100 \times 100$, and $ 125 \times 125$. We evaluate against SAC and PatchZero's original BPDA attack ($L_{e2e}$ and 200 steps of size 0.01 for SAC, 100 steps of 0.01 for PatchZero), SAC and PatchZero's original BPDA attack but with 1000 steps of size 0.002, ($L_{e2e}$ and 1000 steps), and our adaptive attack ($L_{obj\_det} + \lambda L_{patch\_seg}$ and 1000 steps), with our adaptive attack being the most effective.}\label{tab:existing_defenses_box}
\scalebox{0.85}{\begin{tabular}{@{}llc|ccc|ccc|ccc@{}}
\toprule
\multicolumn{1}{c}{\multirow{2}{*}{\textbf{Dataset}}} & \multicolumn{1}{c}{\multirow{2}{*}{\textbf{Def.}}} & \multirow{2}{*}{\textbf{Clean}} & \multicolumn{3}{c|}{\textbf{\begin{tabular}[c]{@{}c@{}}$L_{e2e}$\\ 
$\leq 200$ Steps\end{tabular}}}                 & \multicolumn{3}{c|}{\textbf{\begin{tabular}[c]{@{}c@{}}$L_{e2e}$\\ 1000 Steps\end{tabular}}} & \multicolumn{3}{c}{\textbf{\begin{tabular}[c]{@{}c@{}}$L_{obj\_det} + \lambda L_{patch\_seg}$\\ 1000 Steps\end{tabular}}}                    \\ \cmidrule(l){4-12} 
\multicolumn{1}{c}{}                                  & \multicolumn{1}{c}{}                                  &                                     & \textbf{75} & \textbf{100} & \textbf{125} & \textbf{75}              & \textbf{100}             & \textbf{125}             & \textbf{75} & \textbf{100} & \textbf{125} \\ \midrule
\multicolumn{12}{c}{\textbf{Box mAP}}\\\midrule
\textbf{COCO}                                         & \textbf{SAC}                                          & 0.398                               & 0.313            & 0.335              & 0.280              & 0.201                         & 0.163                          & 0.178                          & \textbf{0.079}            & \textbf{0.038}              & \textbf{0.037}              \\ \midrule
\multirow{2}{*}{\textbf{VOC}}                         & \textbf{SAC}                                          & 0.482                               & 0.436            & 0.389              & 0.341              & 0.201                         & 0.235                          & 0.199                          & \textbf{0.065}            & \textbf{0.008}              & \textbf{0.020}              \\
                                                      & \textbf{PZ}                                           & 0.482                               & 0.442            & 0.392              & 0.353              & 0.425                         & 0.416                          & 0.380                          & \textbf{0.065}          & \textbf{0.040}              & \textbf{0.020}              \\ 
                                                      \midrule
                                                      \multicolumn{12}{c}{\textbf{mAP @ IoU = 0.5}}\\\midrule
                                                      \textbf{COCO}                                         & \textbf{SAC}                                          & 0.618                               & 0.537            & 0.536              & 0.457              & 0.308                         & 0.257                          & 0.288                          & \textbf{0.124}            & \textbf{0.062}              & \textbf{0.062}              \\ \midrule
\multirow{2}{*}{\textbf{VOC}}                         & \textbf{SAC}                                          & 0.767                               & 0.705            & 0.662              & 0.610              & 0.308                         & 0.389                          & 0.358                          & 0\textbf{.116}            & \textbf{0.016}              & \textbf{0.043}              \\
                                                      & \textbf{PZ}                                           & 0.767                               & 0.727            & 0.689              & 0.645              & 0.703                         & 0.709                          & 0.664                          & \textbf{0.118}            & \textbf{0.070}              & \textbf{0.039}              \\ \bottomrule
\end{tabular}}
\end{table}

\section{\defense Defense based on Semantic Priors}\label{ours} 
In this section, given the noticeable semantic patterns in the adaptive attacks presented in Section~\ref{existing}, we propose two new OOD Oracles based on the color properties of the training set. We design these oracles with several properties in mind. Specifically, we tie them to semantic properties such that any attacker trying to evade them will have to avoid certain attack patterns in an understandable way. We also design these to be adaptable - they are tuned on statistics in the natural data so they are quick to tune on new datasets without expensive adversarial training and they can handle differing numbers, sizes, and shapes of patches automatically.

\subsection{Color Histograms Oracle}
The first oracle we propose is the Color Histograms Oracle. Our hypothesis with this oracle is that unusual colors that rarely occur in the natural training set are indicative of OOD pixels. Thus, for any color that appears much more often than normal, pixels of this color are considered OOD.

Formally, let $H_x$ refer to a color histogram with $B \times B \times B$ bins for the image $x$. Let $H_{mean}$ refer to the mean color histogram with $B \times B \times B$ bins over the training dataset. Then, let $\hat{M}_x$ be a $B \times B \times B$ mask over the histogram space such that:
$$\hat{M}_x(a, b, c) = \begin{cases}
1 &\text{if $H_x(a, b, c) > \mu_{hist, (a,b,c)} \wedge H_{mean}(a, b, c) > \nu_{hist}$}\\
0 &\text{otherwise}
\end{cases}$$

where $\mu_{a,b,c}$ and $\nu_{hist}$ are some thresholds. The intuition here is that the first term selects colors that occur more often in the image $x$ than an average image in the dataset, while the second term preserves colors that appear more than $\nu_{hist}$ on average to prevent a naturally significant color from being removed. Finally, we output $M_x$, where $M$ is defined as follows:

$$M_x(h, w) = \hat{M}_x\left(\floor*{\frac{x(h,w)}{B}}\right)$$

In practice, we set $\mu_{hist, (a,b,c)}$ to be some number of standard deviations above the mean number of pixels in bin $(a, b, c)$ over the training set images.

\subsection{HSV Oracle}
The second oracle we propose is the HSV Oracle. Our hypothesis with this oracle is that local regions of pixels with highly saturated colors of many different hues are indicative of OOD pixels. Thus, apply a sliding window operation and mark regions with high saturation and a high number of hues as OOD. 

Formally, let $W_{x, h, w}$ refer to the sliding window of size $t$ centered at pixel $(h, w)$ in the hue channel in the HSV representation of the image $x$. Then, let $x_{h, w, s}$ refer to the saturation of pixel $x_{h,w}$ and let $B$ be the number of hue bins. Then, we output $M_x$, which is defined as follows:

$$M_x(h,w) = \begin{cases}
1 &\text{if $\Big\lvert \Big\{ \floor*{\frac{W_{x,h,w}(i,j)}{B}} \Big\}_{0\leq i\leq t, 0 \leq j \leq t} \Big\rvert > \mu_{hsv} \wedge x_{h,w,s} > \nu_{hsv}$}\\
0 &\text{otherwise}
\end{cases}$$

\subsection{Putting \defense together}
To form our final \defense defense, we combine both the Color Histograms Oracle and the HSV Oracle as a composed function Oracle and we use a masking operation to black out all of the OOD labeled pixels as the OOD Smoother.

\subsection{Attacking \defense}
To adaptively attack \defense, we use the same strategy proposed in Section~\ref{existing}, using two loss functions to independently attack the patch segmenter and the downstream detector. However, \defense's oracles are not differentiable. Thus, to differentiate the patch segmenter loss, we propose to train U-Nets~\cite{ronneberger2015u} to learn to approximate the Color Histograms Oracle and the HSV Oracle independently, and then use this as the approximation in BPDA~\cite{athalye2018obfuscated}. Note that we still use two separate loss functions per the strategy laid out in Section~\ref{existing}, rather than performing an end-to-end attack.

\section{Experiments}\label{experiments}
In this section, we evaluate the effectiveness of \defense against adversarial patch attacks. We measure against mAP @ IoU = 0.5, the primary metric in SAC~\cite{liu2022segment} and PatchZero~\cite{xu2023patchzero}, and COCO~\cite{lin2014microsoft} style 101 point box mAP. We find that \defense increases the mAP @ IoU = 0.5 by up to 40\% over SAC~\cite{liu2022segment} and PatchZero~\cite{xu2023patchzero}. 

\begin{table}[]
\centering
\vspace{-0.25cm}
\caption{Evaluating SAC~\cite{liu2022segment}, PatchZero~\cite{xu2023patchzero}, and \defense on COCO~\cite{lin2014microsoft} style box mAP and mAP @ IoU = 0.5 against adversarial patch attacks over patch sizes of $ 75 \times 75$, $ 100 \times 100$, and $ 125 \times 125$ for MS COCO~\cite{lin2014microsoft} and Pascal VOC~\cite{everingham2010pascal}. We find that using \defense and combining \defense with SAC improves adversarial mAP. }\label{tab:robustness_box}
\begin{tabular}{@{}llc|ccc@{}}
\toprule
\multicolumn{1}{c}{\multirow{2}{*}{\textbf{Dataset}}} & \multicolumn{1}{c}{\multirow{2}{*}{\textbf{Defense}}} & \multirow{2}{*}{\textbf{Clean}} & \multicolumn{3}{c}{\textbf{Adaptive Attack}}                                                                              \\ \cmidrule(l){4-6} 
\multicolumn{1}{c}{}                                  & \multicolumn{1}{c}{}                                  &                                 & \textbf{75 x 75} & \textbf{100 x 100} & \multicolumn{1}{c}{\textbf{125 x 125}} \\ \midrule
\multicolumn{6}{c}{\textbf{Box mAP}}\\\midrule
\textbf{COCO}                                         & \textbf{SAC}                                          & 0.398                           & 0.079            & 0.038              & \multicolumn{1}{c}{0.037}                                                    \\
                                                      & \textbf{OURS}                                         & 0.366                           & 0.225            & 0.143              & \multicolumn{1}{c}{0.137}                                                    \\
                                                      & \textbf{OURS + SAC}                                   & 0.365                           & \textbf{0.298}            & \textbf{0.283}              & \multicolumn{1}{c}{\textbf{0.299}}                                                    \\ \cmidrule(r){1-6}
\textbf{VOC}                                          & \textbf{SAC}                                          & 0.482                           & 0.065            & 0.008              & \multicolumn{1}{c}{0.020}                                                \\
\textbf{}                                             & \textbf{PZ}                                           & 0.482                           & 0.065            & 0.040              & \multicolumn{1}{c}{0.020}                                                   \\
                                                      & \textbf{OURS}                                         & 0.473                           & 0.336            & 0.256              & \multicolumn{1}{c}{0.256}                                                   \\
                                                      & \textbf{OURS + SAC}                                   & 0.473                           & \textbf{0.427}            & \textbf{0.388}              & \multicolumn{1}{c}{\textbf{0.343}} \\ \midrule
                                                      
                                                      \multicolumn{6}{c}{\textbf{mAP @ IoU = 0.5}}\\\midrule

                                                      \textbf{COCO}                                         & \textbf{SAC}                                          & 0.618                           & 0.124            & 0.062              & \multicolumn{1}{c}{0.062}                                                  \\
                                                      & \textbf{OURS}                                         & 0.569                           & 0.342            & 0.224              & \multicolumn{1}{c}{0.216}                                                   \\
                                                      & \textbf{OURS + SAC}                                   & 0.568                           & \textbf{0.469}            & \textbf{0.449}              & \multicolumn{1}{c}{\textbf{0.473}}                                                \\ \midrule
\textbf{VOC}                                          & \textbf{SAC}                                          & 0.767                           & 0.116            & 0.016              & \multicolumn{1}{c}{0.043}                                                 \\
\textbf{}                                             & \textbf{PZ}                                           & 0.767                           & 0.118            & 0.070              & \multicolumn{1}{c}{0.039}                                                  \\
                                                      & \textbf{OURS}                                         & 0.760                           & 0.558            & 0.421              & \multicolumn{1}{c}{0.421}                                                    \\
                                                      & \textbf{OURS + SAC}                                   & 0.760                           & \textbf{0.715}            & \textbf{0.683}              & \multicolumn{1}{c}{\textbf{0.626}}                                                    \\                            
 \bottomrule
\end{tabular}
\end{table}

\subsection{Experimental Setup}
\noindent{\bf Datasets and Models.} We evaluate on the MS COCO~\cite{lin2014microsoft} dataset, Pascal VOC~\cite{everingham2010pascal} dataset, and the CARLA~\cite{dosovitskiy2017carla} simulated dataset available from Armory\footnote{https://github.com/twosixlabs/armory}. For the downstream object detector, we use a Faster RCNN~\cite{ren2015faster} with R50~\cite{he2016deep} backbone for all datasets. For MS COCO, we use the pretrained model available in torchvision~\cite{torchvision2016}. For Pascal VOC, we train a model for 20 epochs with SGD at a learning rate of 0.01 for 10 epochs, 0.001 for 5 epochs, and 0.0001 for 5 epochs, a weight decay of 2e-4, and momentum of 0.9. Following PatchZero~\cite{xu2023patchzero}, we train with the training and validation sets from VOC2007 and VOC2012 and test on the test data from VOC2007. For CARLA, we use the pretrained model from Armory, and test on the test hallucination and test disappearance data splits.

~\\
\noindent{\bf Attack Details.} We test against Mask PGD~\cite{madry2017towards} attacks. For existing attacks from prior works, we adopt the same hyperparameters as the original paper. For our adaptive attacks and for our 1K versions of the existing attacks, we use 1000 steps at a step size of 0.002.

~\\
\noindent{\bf Defense Details.} For SAC~\cite{liu2022segment} and PatchZero~\cite{xu2023patchzero}, we use the default settings from the original papers. For \defense, we adapt the parameters to limit the number of pixels changed by \defense on natural images. For MS COCO~\cite{lin2014microsoft}, we set $B = 16$, $\mu_{hist,(a,b,c)}$ to 6 standard deviations above the average histogram value of bin $(a,b,c)$ in the training set, $\nu_{hist} = 10$ $t=9$, $\mu_{hsv}=11$, $\nu_{hsv}= 60 / 255$. For Pascal VOC~\cite{everingham2010pascal}, we set $B = 16$, $\mu_{hist,(a,b,c)}$ to 8 standard deviations above the average histogram value of bin $(a,b,c)$ in the training set, $\nu_{hist} = 5$ $t=9$, $\mu_{hsv}=11$, $\nu_{hsv}= 60 / 255$. For CARLA~\cite{dosovitskiy2017carla}, we use $B = 16$, $\mu_{hist,(a,b,c)}$ to 4 standard deviations above the average histogram value of bin $(a,b,c)$ in the training set, $\nu_{hist} = 25$ $t=9$, $\mu_{hsv}=11$, $\nu_{hsv}= 60 / 255$.

For the differentiable approximations for \defense, we train the same U-Net~\cite{ronneberger2015u} architecture as in SAC~\cite{liu2022segment}. We train on 30\% natural images and 70\% modified images, where the modified images replace a randomly selected $100 \times 100$ patch with random colors. The U-Net is then trained with the output mask of the corresponding \defense OOD oracle as the label, to mimic the behavior of the oracle. We train for 5 epochs with Adam at a learning rate of 1e-4 and a weight decay of 1e-5.

\subsection{Robustness Evaluation}

We include results for attacking \defense in Table~\ref{tab:robustness_box}  on COCO~\cite{lin2014microsoft} style 101 points interpolated box mAP and mAP @ IoU = 0.5 for MS COCO~\cite{lin2014microsoft} and Pascal VOC~\cite{everingham2010pascal}. Results for CARLA~\cite{dosovitskiy2017carla} are in Table~\ref{tab:carla}. As before, following SAC~\cite{liu2022segment}, we evaluate on 1000 images. We test each defense against our adaptive attack with 1000 steps, which we showed in Section~\ref{existing} to be the most effective against SAC~\cite{liu2022segment} and PatchZero~\cite{xu2023patchzero}. 
\begin{table}[]
\centering
\caption{Evaluating SAC~\cite{liu2022segment}, PatchZero~\cite{xu2023patchzero}, and \defense on COCO~\cite{lin2014microsoft} style box mAP and mAP @ IoU = 0.5 on CARLA~\cite{dosovitskiy2017carla}. Since predefined masks are provided, we attack those masks rather than squares.}\label{tab:carla}
\begin{tabular}{@{}llcc@{}}
\toprule
\multicolumn{1}{c}{\textbf{Dataset}} & \multicolumn{1}{c}{\textbf{Defense}} & \textbf{Clean}             & \textbf{Adaptive Attack} \\ \midrule
\multicolumn{4}{c}{\textbf{Box mAP}}                                                                                                \\ \midrule
\multirow{4}{*}{\textbf{CARLA}}      & \textbf{SAC}                         & \multicolumn{1}{c|}{0.226} & 0.020                    \\
                                     & \textbf{PZ}                          & \multicolumn{1}{c|}{0.226} & 0.094                    \\
                                     & \textbf{OURS}                        & \multicolumn{1}{c|}{0.195} & \textbf{0.195}           \\
                                     & \textbf{OURS + SAC}                  & \multicolumn{1}{c|}{0.195} & 0.186                    \\ \midrule
\multicolumn{4}{c}{\textbf{mAP @ IoU = 0.5}}                                                                                        \\ \midrule
\multirow{4}{*}{\textbf{CARLA}}      & \textbf{SAC}                         & \multicolumn{1}{c|}{0.353} & 0.035                    \\
                                     & \textbf{PZ}                          & \multicolumn{1}{c|}{0.353} & 0.156                    \\
                                     & \textbf{OURS}                        & \multicolumn{1}{c|}{0.316} & \textbf{0.316}           \\
                                     & \textbf{OURS + SAC}                  & 0.316                      & 0.304                    \\ \bottomrule
\end{tabular}
\end{table}

For both metrics on all datasets, we find that \defense is more robust on all patch sizes. We see a 10\% gain on MS COCO for box mAP, a 20\% gain on MS COCO for mAP @ IoU = 0.5 and VOC for box mAP, and a 35\% gain on VOC for mAP @ IoU = 0.5. For MS COCO~\cite{lin2014microsoft} and Pascal VOC~\cite{everingham2010pascal}, we find that we can further increase the robustness by combining \defense's oracles with SAC~\cite{liu2022segment}, adding at least another 7\% and up to 26\% across settings.

\subsection{Transfer Attacks}
To test how well SAC~\cite{liu2022segment} and PatchZero~\cite{xu2023patchzero} capture the semantic priors introduced by \defense, we test how well our adaptive attacks generated on each of these three defenses transfers to each other. We show the results for Pascal VOC~\cite{everingham2010pascal} on patch attacks of size $100 \times 100$ in Table~\ref{tab:transfer}. We find that \defense is much more robust on these attacks than SAC~\cite{liu2022segment} or PatchZero~\cite{xu2023patchzero} were. Given that the only difference is the OOD Oracle, this suggests that the ML-based systems in SAC and PatchZero struggled to pick up on the color-based semantic priors encoded by \defense.

\begin{table}
\vspace{-0.5cm}
\centering
\caption{We find that \defense maintains higher adversarial mAP against transfer attacks generated from SAC~\cite{liu2022segment} and PatchZero~\cite{xu2023patchzero}, suggesting that SAC and PatchZero failed to learn the semantic priors encoded in \defense. Metric: mAP @ IoU = 0.5.}\label{tab:transfer}
\begin{tabular}{@{}lccc@{}}
\toprule
                            \diagbox[dir=NW]{Attack}{Defense}           & \multicolumn{1}{c}{\textbf{SAC}}   & \multicolumn{1}{c}{\textbf{PZ}} & \multicolumn{1}{c}{\textbf{OURS}} \\ \midrule
\multicolumn{1}{l}{\textbf{SAC}}       & \multicolumn{1}{c}{0.016} & \multicolumn{1}{c}{0.170}     & 0.514                     \\
\multicolumn{1}{l}{\textbf{PZ}} & \multicolumn{1}{c}{0.053} & \multicolumn{1}{c}{0.070}     & 0.41                      \\
\multicolumn{1}{l}{\textbf{OURS}}      & \multicolumn{1}{c}{0.193} & \multicolumn{1}{c}{0.168}     & 0.421                     \\ \bottomrule
\end{tabular}
\vspace{-0.5cm}

\end{table}


\section{Discussion}
We next discuss: 1) generalizability of \framework, 2) adaptability of \defense, 3) generative attacks, 4) failure modes and limitations, and 5) societal impact.

~\\
\noindent{\bf Generalizing \framework: }
We note that \framework could be extended beyond simply patch attacks on object detectors. Theoretically, \framework could capture a variety of definitions of OOD and smoothers could then take on different forms to satisfy different properties based on the threat model and task. We leave the exploration of this direction to future work.

~\\
\noindent{\bf Adaptability of \defense: }
\defense has a key benefit that makes adapting it to new datasets more practical than SAC~\cite{liu2022segment} or PatchZero~\cite{xu2023patchzero}. Specifically, it does not require an expensive adversarial training loop with a multi-step adversary. This was particularly limiting in the case of PatchZero, which took approximately 25 hours per epoch for its stage 2 training on Pascal VOC~\cite{everingham2010pascal} and 21 hours for CARLA~\cite{dosovitskiy2017carla} on a machine with 2 RTX 3090 GPUs. Training on MS COCO~\cite{lin2014microsoft} would have taken even longer, considering Pascal VOC's 2007 and 2012 training and validation sets have a combined 16551 images while MS COCO has 118287 images. CARLA had only 3600 images but were at much higher resolution, making it difficult to fit in GPUs with any memory smaller than a 3090. 

~\\
\noindent{\bf Generative Attacks:}
In this work, we focus specifically on gradient-based attacks. One line of attacks we did not consider are diffusion generated attacks~\cite{lin2023diffusion}. Such attacks likely follow different OOD distributions and may require different OOD oracles. However, \framework is flexible and could be extended to include OOD oracles that target such generative attacks.

~\\
\noindent{\bf Failure Modes and Limitations:}
One limitation of \defense is that if any natural objects rely on rare occasions the use of unnatural colors targeted by the color oracles, that object could then disappear. Also, images that are highly concentrated in one color are more likely to have false positive pixels. In addition, following SAC~\cite{liu2022segment}, in \defense we simply removed the OOD features by setting such pixels to black. However, it may be possible to reconstruct something close to the natural image under the patch by using a high-quality inpainter such as DDNM~\cite{wang2022zero}. We leave the exploration of this direction to future work. 

~\\
\noindent{\bf Societal Impact:}
Our results show that basic ML-based patch detection systems should not be trusted to protect against such threats. \defense is a step taken in a direction aimed at improving the robustness of object detectors against patch attacks, which is imperative for the safe deployment of such systems. We will release the code publicly upon acceptance so that other researchers can build on our work (code is also included in the supplementary material).
\section{Conclusion}
In this paper we unify existing object detection patch attack defenses under a general framework called \framework. \framework guides us towards the development of a novel attack that breaks existing defenses in SAC~\cite{liu2022segment} and PatchZero~\cite{xu2023patchzero}. Our key insight is that ML models struggle to learn semantic priors without explicit supervision. We thus propose \defense, a defense that explicitly incorporates semantic priors, raising the adversarial robustness by up to 40\% alone and up to 60\% when combined with SAC~\cite{liu2022segment}. 

\section{Acknowledgements}
This material is based on work supported by DARPA under agreement number 885000, the National
Science Foundation (NSF) Grants 2039445, CCF-FMitF-1836978, SaTC-Frontiers-1804648 and CCF1652140, Air Force Grant
FA9550-18-1-0166, ARO grant number W911NF-17-1-0405, and National Science Foundation Graduate Research
Fellowship Grant No. DGE 1841052. Feng is partially supported by the J. Robert Beyster Computational Innovation Graduate Fellowship. The U.S. Government is authorized to reproduce
and distribute reprints for Governmental purposes notwithstanding any copyright notation thereon.
Any opinions, findings, and conclusions or recommendations expressed in this material are those of the
author(s) and do not necessarily reflect the views of our research sponsors.

%
%
\bibliographystyle{splncs04}
\bibliography{main}
\end{document}